%% file: main.tex
\documentclass[10pt,twocolumn,letterpaper]{article}

\usepackage{cvpr}              
\usepackage{mathtools}
\usepackage[accsupp]{axessibility}

\input{preamble}

\definecolor{cvprblue}{rgb}{0.21,0.49,0.74}
\usepackage[pagebackref,breaklinks,colorlinks,citecolor=cvprblue]{hyperref}

\def\paperID{7956} 
\def\confName{CVPR}
\def\confYear{2024}

\title{Watermark-embedded Adversarial Examples for \\ Copyright Protection
against Diffusion Models}

\author{Peifei Zhu, Tsubasa Takahashi, Hirokatsu Kataoka  \\
LY Corporation \\
{\tt\small \{peifei.zhu, tsubasa.takahashi, jpz4219, \}@lycorp.co.jp}}

\begin{document}
\maketitle
\input{sec/0_abstract}    
\input{sec/1_intro}

\input{sec/2_relatedworks}

\input{sec/3_AdvWMGen}
\input{sec/4_experiments}

\input{sec/5_conclusions}

{
    \small
    \bibliographystyle{ieeenat_fullname}
    \bibliography{main}
}

\input{sec/X_suppl}

\end{document}

%% file: preamble.tex
\usepackage[dvipsnames]{xcolor}


%% file: sec/0_abstract.tex
\begin{abstract}
Diffusion Models (DMs) have shown remarkable capabilities in various image-generation tasks. However, there are growing concerns that DMs could be used to imitate unauthorized creations and thus raise copyright issues. To address this issue, we propose a novel framework that embeds personal watermarks in the generation of adversarial examples. Such examples can force DMs to generate images with visible watermarks and prevent DMs from imitating unauthorized images. We construct a generator based on conditional adversarial networks and design three losses (adversarial loss, GAN loss, and perturbation loss) to generate adversarial examples that have subtle perturbation but can effectively attack DMs to prevent copyright violations. Training a generator for a personal watermark by our method only requires 5-10 samples within 2-3 minutes, and once the generator is trained, it can generate adversarial examples with that watermark significantly fast (0.2s per image). We conduct extensive experiments in various conditional image-generation scenarios. Compared to existing methods that generate images with chaotic textures, our method adds visible watermarks on the generated images, which is a more straightforward way to indicate copyright violations. We also observe that our adversarial examples exhibit good transferability across unknown generative models. Therefore, this work provides a simple yet powerful way to protect copyright from DM-based imitation.
\end{abstract}

%% file: sec/1_intro.tex
\section{Introduction}
\label{sec:intro}

Diffusion models (DMs) \cite{ho2020denoising, song2020score, rombach2022high} have garnered significant attention in both the computer vision industry and academia. Compared to previous generative models such as generative adversarial networks (GANs) \cite{goodfellow2014generative, mirza2014conditional, karras2019style}, DMs have demonstrated a remarkable ability to generate realistic images with higher resolution and diversity. Despite the positive aspects of DMs, there are public concerns about potential copyright violations when unauthorized images are used to train DMs or to generate new creations by DMs \cite{yang2022diffusion}. For example, an entity could use copyrighted paintings shared on the Internet to generate images with a similar style and appearance, potentially earning illegal revenue. Therefore, it is necessary to develop techniques to prevent the imitation of unauthorized creations by DMs.

\begin{figure}[t]
   \begin{center}
      \includegraphics[width=1.0\linewidth]{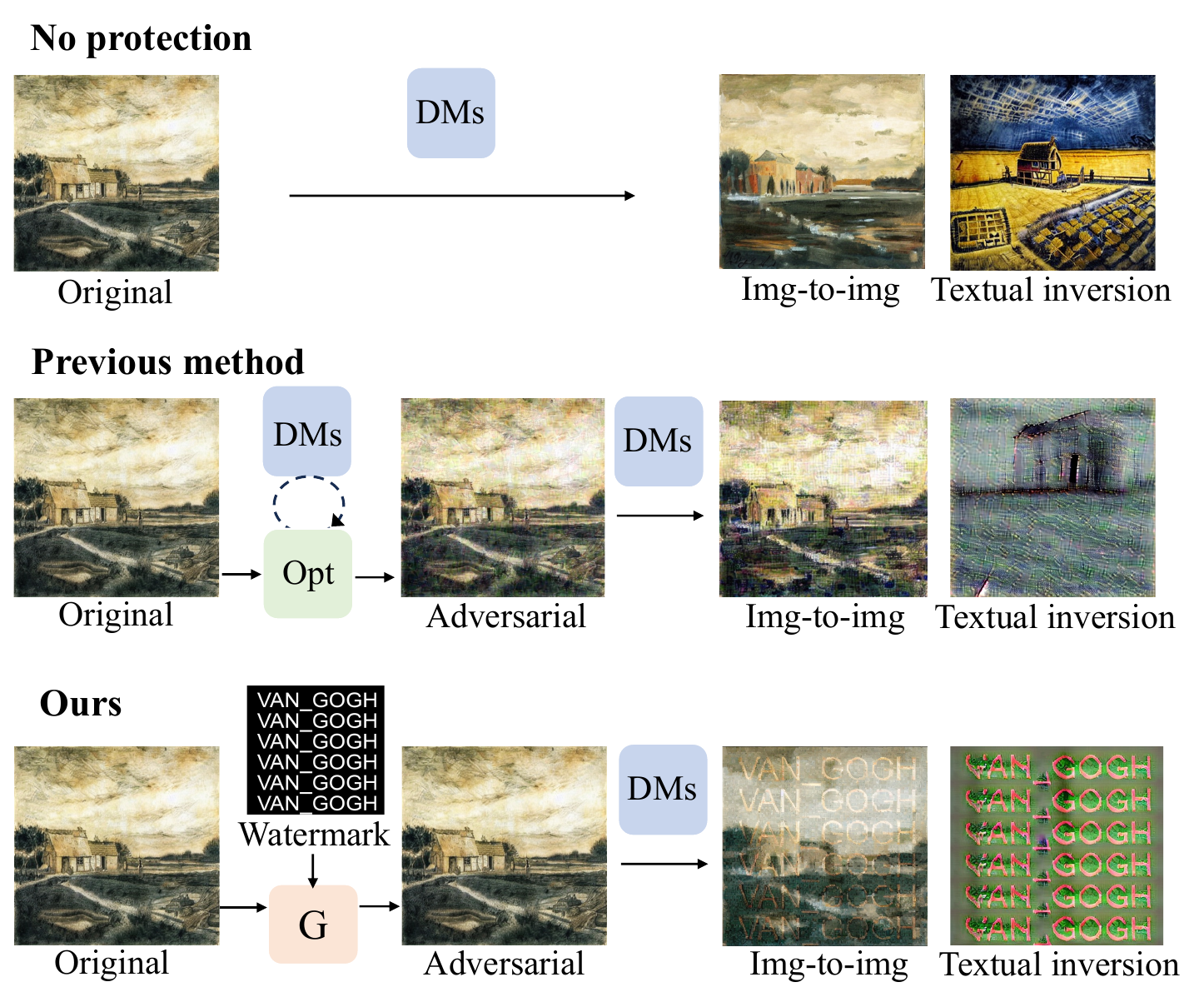}
   \end{center}
      \caption{Copyright issues of DMs and adversarial example-based methods for copyright protection. Without protection, DMs can easily imitate original images under different image generation scenarios. A previous method (AdvDM \cite{liang2023adversarial}) generates adversarial examples by optimization against DMs to prevent DMs from extracting the feature of the original images, resulting in the generation of chaotic images. Our method goes one step further to build a generator that embeds personal watermarks into the generation of adversarial images. Such examples force DMs to generate images with visible watermarks for tracing copyright. Our method is fast, simple yet powerful in protecting copyrights against DMs.}
   \label{fig:advwatermark_overview}
   \end{figure}

Several previous works have been proposed to address these concerns. Watermarking \cite{mohanty1999digital, begum2020digital} is a common solution that embeds an invisible message into the image, which can then be extracted to identify the existence of copyright. Recent techniques typically integrate watermarking into the generative process of DMs \cite{fernandez2023stable, cui2023diffusionshield, peng2023protecting}. However, these methods have several limitations. First, they are primarily designed to protect the copyright of DMs or to distinguish generated images from natural ones, which differs from our goal of protecting the copyright of human creations. Second, embedding watermarks into the generative process requires re-training or at least fine-tuning DMs, and a post-process to extract the watermark is needed to identify it, which can be costly. On the other hand, methods that use adversarial examples for DMs to protect copyright have been proposed recently \cite{liang2023adversarial, liang2023mist, van2023anti}. The idea is to generate adversarial examples to hinder DMs from extracting the features of the original images. This generation process does not require any re-training of DMs or any post-processing for the generated image. However, there are still several limitations. First, even though these adversarial examples could add chaotic textures to the generated images, such changes are difficult to comprehend. For instance, users may wonder if the problem originates from the DM itself rather than considering the copyright issue. Second, the original image's copyright is not traceable since no copyright-related information is embedded. Third, each adversarial example needs to be optimized separately against DMs, which makes the generation time-consuming.

In this work, we propose a simple yet powerful method to protect copyrighted images from imitation by DMs. Different from previous adversarial example-based methods, we go one step further by embedding personal watermarks into the generation of adversarial examples for copyright tracing. Such examples could force DMs to generate images with visible watermarks as well as chaotic textures. In addition, instead of using iterative optimization against DMs to generate examples, we train a generator beforehand using only a few samples. Once the generator is trained, it can be used to generate adversarial examples significantly fast. As shown in Figure \ref{fig:advwatermark_overview}, instead of posting the original image on the Internet, posting adversarial examples with only subtle changes could prevent DMs from imitation. Compared to previous methods, our method provides a more straightforward way to warn of potential copyright violations.

Our contributions can be summarized as follows:
\setlist{nolistsep} 
\begin{itemize}
    \item We design a novel framework that embeds personal watermarks into the generation of adversarial examples to prevent copyright violations caused by DM-based imitation. Our adversarial examples can force DMs to generate images with visible watermarks for tracing copyright.
   \item We train a generator based on a conditional GAN architecture to generate adversarial examples (Figure \ref{fig:architecture}). We design three losses: adversarial loss, GAN loss, and weighted perturbation loss. These losses aim to improve the attack ability of the adversarial examples while making the perturbation invisible to the human eye.
   \item We conduct extensive experiments in various image generation scenarios, along with robustness evaluations against defenses and assessments of transferability to other models. The experiments demonstrate that our adversarial examples can prevent unauthorized images from being learned and can generate visible watermarks for copyright tracing. Our generation process is significantly fast (0.2s per image), and the generated examples also exhibit good transferability across other generative models.
\end{itemize}

%% file: sec/2_relatedworks.tex
\section{Related Works}
\label{sec:related}

\subsection{Generative Diffusion Models}
In recent years, DMs have made significant advancements in various fields \cite{sohl2015deep, dhariwal2021diffusion, nichol2021improved, yang2022diffusion}. DMs are a type of generative model that simulates a diffusion process to generate new data samples. DMs consist of a forward process and a reverse process. The forward process involves gradually adding Gaussian noise to the data over a series of time steps until it becomes completely random. The reverse process gradually removes noise through a series of learned denoising steps, reconstructing the data back to its original form or generating new samples from the same distribution.

DMs have been applied to various generation tasks such as image synthesis \cite{ruiz2023dreambooth, dhariwal2021diffusion}, text-to-image generation \cite{balaji2022ediffi, ramesh2022hierarchical, zhang2023adding, zhang2023text}, and text-guided image editing \cite{choi2023custom, zhang2023sine, yang2023paint, mokady2023null}. Among these, latent diffusion models \cite{rombach2022high}, which perform the diffusion steps in the latent image space, have demonstrated a remarkable ability to generate high-resolution and realistic images. Despite the impressive performance, some concerns that generating new creations by DMs based on unauthorized images could lead to copyright issues.

\subsection{Image Watermarking}
Image watermarking is a technique that embeds a piece of information into an image to protect the image from copyright infringement. Traditional watermarking methods typically involve altering pixel values \cite{cox2002digital} or manipulating the frequency domain of an image \cite{navas2008dwt} to embed watermarks. More recently, deep-learning-based methods have been developed. A common approach is using encoder-extractor networks, where an encoder embeds the watermark into the image, and an extractor is trained to retrieve the watermark from the marked image \cite{zhu2018hidden, zhang2019invisible, tancik2020stegastamp}.

In terms of watermarking for generative models, most existing works focus on protecting the copyright of generative models \cite{fei2022supervised, wu2020watermarking, liu2023watermarking, yu2021artificial, peng2023protecting} or distinguishing generative images from natural images \cite{yu2020responsible, fernandez2023stable, zhao2023proactive, cui2023diffusionshield}. These methods involve reformulating the training process of DMs to inject watermarks into the images, as well as developing a decoder to identify the embedded messages from the generated images. However, the re-training can be costly and it is impractical to check every generated image with a decoder.

\subsection{Adversarial Examples for Generative models}
Adversarial examples are created by adding small but intentional perturbations to input samples such that the perturbed inputs cause the model to produce incorrect answers. Various adversarial attack methods have been proposed to confuse classification models \cite{goodfellow2014explaining, madry2017towards, xiao2018generating, dong2018boosting} or segmentation models \cite{xie2017adversarial, zhu2023frequency, fischer2017adversarial}. There are also several existing works studying adversarial examples for generative models such as GANs \cite{kos2018adversarial}, variational autoencoders (VAEs) \cite{tabacof2016adversarial}, and flow-based generative models \cite{pope2020adversarial}.

Recently, methods that generate adversarial examples for DMs \cite{liang2023adversarial, liang2023mist, van2023anti} have been proposed. These methods maximize the diffusion loss to obtain a perturbation, and adding such a perturbation to the original image enables DMs to generate images with chaotic content. However, these methods have several drawbacks: 1) such chaotic content is not straightforward to indicate copyright violation, 2) the copyright of the original image is not traceable, and 3) optimization against DMs is time-consuming. We propose a simple yet powerful method that directly incorporates personal watermarks into the generation of adversarial examples, and we build a generator that can generate adversarial examples significantly faster than existing methods.

%% file: sec/3_AdvWMGen.tex
\section{Proposed Method}

\subsection{Problem Statement}
Given an original image $x$, we have a diffusion model $\theta$ that can learn the data distribution of $x$ and generate images with similar style and appearance. In this work, we aim to generate an adversarial example $x^\prime$ that can prevent $\theta$ from extracting the feature of $x$ and can directly show copyright information on the generated image. For example, for image-to-image generation, the generated image from $x^\prime$ by using $\theta$ can be denoted as $\theta\left(x^\prime\right)$. We design a way to provide the copyright information by using a watermark that contains the ownership details (e.g. artist name) of the original image. Therefore, given a watermark image $m$, the adversarial example $x^\prime$ that embeds $m$ can be calculated by
\begin{equation}
\begin{split}
    x^\prime & \coloneqq \underset{x^\prime}{\mathrm{argmin}} \|{\theta \left(x^\prime \right) - m}\|, \\
    &\text{s.t.} \quad \| x - x^\prime \| \leq \sigma,
    \label{equ:adv_loss1}
\end{split}
\end{equation}
where $\sigma$ is a constant to control the range of the perturbation. This is a targeted attack setting where we compel the image generated by DMs to be closer to the watermark, while ensuring that the adversarial example remains as close to the original one as possible.

However, several issues are associated with generating adversarial examples using Equation \ref{equ:adv_loss1}. First, directly optimizing $x^\prime$ against DMs for each $x $ could be time-consuming, so we need to find a way to speed up the generation process. Second, since there are several image-generation scenarios related to imitating the original image, such as text-guided image-to-image generation \cite{choi2023custom}, text-to-image generation under textual inversion \cite{gal2022image} or DreamBooth \cite{ruiz2023dreambooth}, we aim to generate an adversarial example $x^\prime$ that can be applicable to all these scenarios. Meanwhile, we also expect that $x^\prime$ could have good transferability on other generative models. Third, since our target attack aims to add personal watermarks to the generated images, the perturbation naturally becomes larger in the watermark regions. We need a solution to make the perturbation invisible to human eyes. 

To address the first issue, we train a generator beforehand instead of implementing optimization every time. To address the second and third issues, we design an adversarial loss, a GAN loss, and a perturbation loss to improve the attack ability across various tasks and models while keeping the perturbation invisible.

\subsection{Architecture Overview}
\label{sec:proposed}
\begin{figure*}[t]
   \begin{center}
      \includegraphics[width=0.83\linewidth]{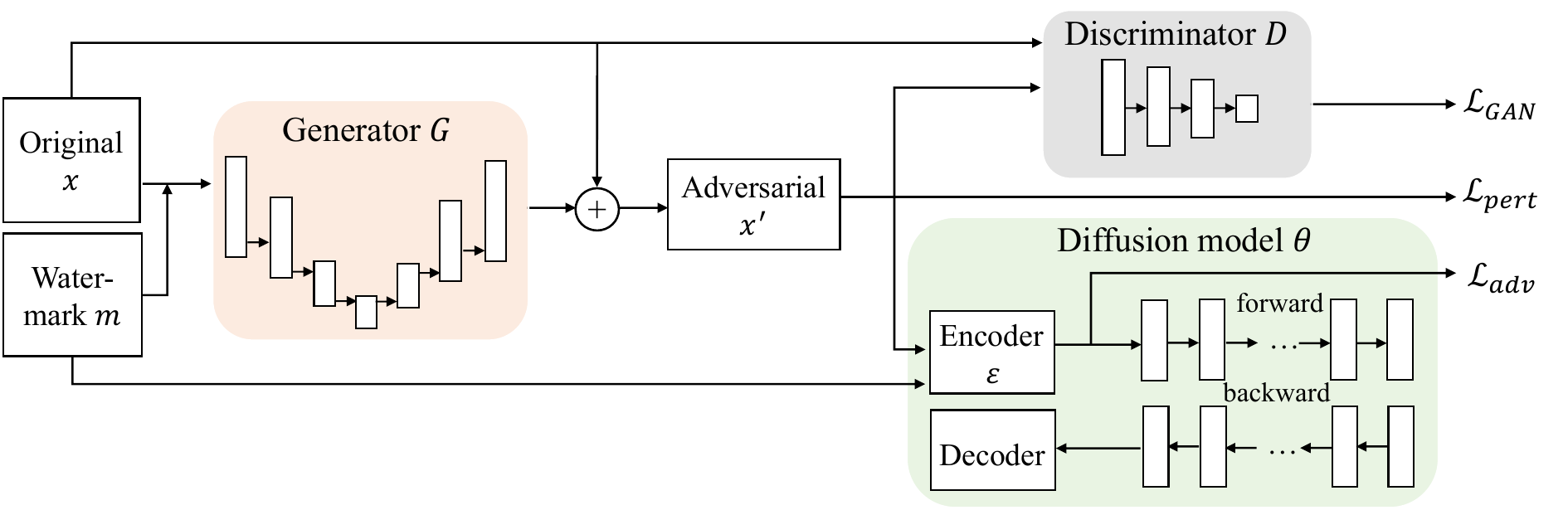}
   \end{center}
      \caption{Architecture overview. $G$ generates perturbation for $x$ conditioned on $m$. $D$ and $G$ produce $\mathcal{L}_{GAN}$ to compel $x^\prime$ to be closer to $x$. $\mathcal{L}_{adv}$ aims to force the image generated by $\theta$ to display a visible watermark $m$, and $\mathcal{L}_{pert}$ further bounds the magnitude of the perturbation.}
   \label{fig:architecture}
\end{figure*}

We propose a conditional GAN architecture for generating watermark-embedded adversarial examples. The architecture of our method contains three components: a generator $G$, a discriminator $D$, and a target DM $\theta$, shown in Figure \ref{fig:architecture}. The generator $G$ has an encoder to extract features from the inputs and a decoder to generate perturbation using the features. The inputs of $G$ are the original image $x$ and the watermark $m$, and the perturbation is generated conditioned on $m$. We perform the conditioning by concatenating $x$ and $m$ along the channel dimension and feeding them into $G$. The adversarial example $x^\prime$ can be obtained by adding $x$ and the generated perturbation. The discriminator $D$ is a classifier that distinguishes $x^\prime$ from $x$, and the target DM $\theta$ is always kept frozen. 

We design three losses to optimize $G$ and $D$. 1) A \textbf{GAN loss} aims to compel $x^\prime$ to be closer to $x$. By using the GAN loss, we can train $G$ to produce images that $D$ cannot distinguish from input images. 2) A \textbf{perturbation loss} is designed to bound the magnitude of the perturbation and make it invisible. 3) An \textbf{adversarial loss} is used for $G$ to generate perturbation that can attack DMs under various scenarios. The details are described in Section \ref{sec:loss_function}. Once the generator $G$ is optimized, it can be used directly to generate adversarial examples significantly fast.

\subsection{Loss functions}
\label{sec:loss_function}
\noindent \textbf{GAN loss.} $\mathcal{L}_{GAN}$ is used to quantify the difference between the adversarial example and the original image. For a set of images, $\mathcal{L}_{GAN}$ can be written as:
\begin{equation}
   \mathcal{L}_{GAN}=\mathbb{E}_x\log{D\left(x\right)}+\mathbb{E}_x\log{\left\{1-D\left(x^\prime\right)\right\}}.
\end{equation}
The adversarial image $x^\prime$ is generated under the condition of the watermark $m$, which can be written as:
\begin{equation}
    x^\prime=x+G\left(x \mid m\right).
\end{equation}

\noindent \textbf{Perturbation loss.} $\mathcal{L}_{pert}$ aims to limit the magnitude of the perturbation. We design this loss based on the soft hinge loss used in previous adversarial attacks \cite{carlini2017towards, liu2016delving, xiao2018generating}. However, the soft hinge loss assigns equal importance to all perturbations within the image. In our case, the perturbation in the watermark region will be larger than that in other regions, making the watermark visible even in the adversarial example. To better conceal the watermark as well as bound the perturbation in the whole image, we design a weighted perturbation loss, where we assign larger weights $w$ to the watermark region. This encourages the perturbation in the watermark region to be smaller and invisible. For a set of images, this loss can be expressed as:
\begin{equation}
   \mathcal{L}_{pert}=\mathbb{E}_x \max (0, \left \|G\left(x \mid m\right) (1+w \cdot m)\|_2 - c \right),
\end{equation}
where $c$ is a constant denoting a user-specified bound.

\noindent \textbf{Adversarial loss.} $\mathcal{L}_{adv}$ is used to generate a perturbation that can attack DMs. As we expect our adversarial examples to have transferability across different image generation scenarios and various generative models, we design the loss based on the latent representation of latent diffusion models (LDM). We choose LDM for two reasons. First, LDM has outperformed other models in both high-quality image generation and sampling efficiency, and it has been used in various generative models. Generating adversarial examples against LDM naturally increases their transferability across various models. Second, LDM maps an image $x$ to a latent representation $\varepsilon \left(x\right)$ which is the output of the Variational Auto-Encoder (VAE). This representation is widely used in conditional image generation scenarios, and it provides a simple way to control the image generation of DMs. Therefore, instead of directly exploiting the generated image, we minimize the distance between the representation of the adversarial example $\varepsilon \left(x^\prime\right)$ and that of the watermark $\varepsilon \left(m\right)$, formulated as:

\begin{equation}
    \mathcal{L}_{adv}=\mathbb{E}_x \|\varepsilon \left(x^\prime\right)- \varepsilon \left(m\right)\|_2.
\end{equation}

\noindent \textbf{Optimization.}
Finally, we combine the above three losses to formulate the objective function for training, written as:
\begin{equation}
\min_{G}\max_{D}V(D,G)=\mathcal{L}_{adv}+\alpha\mathcal{L}_{GAN}+\beta\mathcal{L}_{pert},
\label{equation:total}
\end{equation}
where $\alpha$ and $\beta$ are weights to control the balance of the three losses. By playing a minimax game between the $G$ and $D$, the optimal parameters of the model can be obtained.

\subsection{Image Generation under Different Settings}
\label{sec:imagegen_settings}
Since our target is to prevent copyright violations, we focus on conditional image generation scenarios where DMs sample the distribution of the original images to generate new ones. Such scenarios include text-guided image-to-image generation \cite{nichol2021glide} and text-to-image generation under textual inversion \cite{gal2022image} described in this section, and other scenarios including DreamBooth \cite{ruiz2023dreambooth}, LoRA \cite{hu2021lora}, and Custom Diffusion \cite{kumari2023multi} discussed in Appendix \ref{sec:OtherScenarios}.

\noindent \textbf{Text-guide image-to-image generation.}
In this setting, we can pass a text prompt and an input image to condition the image generation. Our method can be directly applied to this setting, as shown in Figure \ref{fig:architecture}. During the training of the generator and discriminator, we simply use the same prompt (e.g. ``A painting") for all samples. For the inference, any prompt can be used for generating new images. 

\noindent \textbf{Textual inversion.}
Textual inversion is a method for capturing new concepts from a small number of images. These new concepts can then be used to control image generation in text-to-image settings. The steps to apply our method in this setting are as follows. First, given a small set of images within a similar category (e.g. same artist), we extract a text $S_*$ to represent our new concept. We then replace the vector associated with the tokenized string with an optimized embedding $v_*$. The optimization can be applied by minimizing the LDM loss over the images from the small set. During the image generation by DMs, the text $S_*$ is used as a condition to generate new images. We conduct this generation for both original images and adversarial images. The adversarial images are generated through the image-to-image generation setting as shown in Figure \ref{fig:architecture}.

%% file: sec/4_experiments.tex
\section{Experiments}
In this section, we evaluate our method for generating adversarial examples for DMs. We conduct experiments under text-guided image-to-image generation and textual inversion, performing various analyses to assess the effectiveness, robustness, and transferability of our method. We perform the evaluation on WikiArt \cite{artgan2018} and ImageNet \cite{imagenet15russakovsky} datasets. All experiments are run on an NVIDIA A100 80GB GPU. Training the generator using 100 samples with 200 epochs takes about 20 minutes.

\subsection{Experimental Settings}

\noindent
\textbf{Datasets.} The WikiArt dataset contains 50k paintings from 195 artists. We randomly select 3000 paintings from 50 artists. We create different watermarks for different artists. These watermarks are binary images with a black background and white artist names such as ``VAN\_GOGH", ``CLAUDE\_MONET". Since only a small number of samples are needed to train our generator, we select 10 images from each artist (500 images in total) for training and the remaining 2500 images are for evaluation. For the ImageNet dataset, we randomly select 2000 images (100 for training, 1900 for evaluation) from the goldfish, tiger shark, peacock, goose, Eskimo dog, and tabby cat category. The watermarks are designed as ``IMAGENET\_CAT", ``IMAGENET\_FISH", ``IMAGENET\_DOG", etc.

\noindent
\textbf{Evaluation metrics.} We evaluate our method from two perspectives. For the adversarial example, we use MSE, PSNR, and SSIM to measure the image quality and the difference from the original image. For the image generated by DMs, we use Fréchet Inception Distance (FID) \cite{heusel2017gans} and precision (prec.) \cite{kynkaanniemi2019improved} to measure the similarity between the generated and original images. Since we expect DMs to generate images with visible watermarks, we use Normalized Cross-Correlation (NCC) \cite{jahne2005digital} to measure the similarity between the generated image and the watermark. Please note that, unlike traditional watermark methods, we do not extract the watermark from the generated image, and thus NCC is smaller in our case. However, to reduce the influence of the background, we first calculate a difference image between the image generated from the original image and that from the adversarial example, and then use this difference image and the watermark to obtain NCC.

\noindent
\textbf{Implementation details.} For the generator $G$, we adopt an architecture similar to that in \cite{johnson2016perceptual}. For the discriminator $D$, we use an architecture similar to the discriminator in AdvGAN \cite{xiao2018generating}. Further details regarding the implementation of our model are provided in Appendix \ref{sec:modeldetails}. We set the weight of GAN loss $\alpha=1.0$, the weight of perturbation loss $\beta=10$, the weight for watermark region $w=4$, and the bound $c=10/255$. During the model training, we set the batch size to 8 and the learning rate to $0.001$. For comparison, we use AdvDM \cite{liang2023adversarial} and Mist (fuse mode) \cite{liang2023mist}, which also use adversarial examples to protect copyright. We set their sampling step to $100$ and the maximum perturbation to $10/255$.

\subsection{Text-guided Image-to-Image Generation}
\label{sec:Text-guided Image-to-Image Generation}
We evaluate the performance of our method under the setting described in Section \ref{sec:imagegen_settings}. In this setting, a strength parameter is used to control the level of noise added to the original image during the generation of new images. A strength value close to 0 produces an image nearly identical to the original, while a value close to 1 results in an image that significantly differs from the original. We set the strength to 0.3 to simulate realistic imitation, and additional results with different strength values can be found in the Appendix \ref{sec:strengthresult}. For the text prompt, we uniformly use ``A painting" for all images for convenience, and experiments using varied prompts are also included in the Appendix \ref{sec:variedprompts}.

We first show the evaluation result of the image generated by LDM based on either the original image or the adversarial example. A high FID and a low precision mean that the generated image is far from the original one, which further indicates that LDM fails to capture the feature of the original image. As shown in Table \ref{table:generation_eval}, our adversarial examples largely increase FID and decrease precision. We also calculate NCC between the generated image and the watermark. For Mist, since it is possible to input a target image into the generation, we apply our idea to use different watermarks as the target images and use the same method to calculate NCC. Results show that our method has a significantly higher NCC, which means our adversarial examples can force LDM to generate images with more visible watermarks. These results suggest our method has great potential to prevent copyright violations caused by LDM imitation.

\begin{table}[t]
   \begin{center}
   \begin{tabular}{ccccc}
   \hline
   Method & NCC $\uparrow$ & FID $\uparrow$ & prec. $\downarrow$ & recall \\ 
   \hline
   Original & 0 & 120.7 & 0.85 & 0.28 \\
   AdvDM  & 0 & 228.2 & \textbf{0.03} & 0.25 \\
   Mist  & 0.09 & 205.8 & 0.04 & 0.33 \\
   Ours  & \textbf{0.31} & \textbf{245.5} & \textbf{0.03} & 0.29 \\
   \hline
   \end{tabular}
   \end{center}
   \caption{The performance of the generated image under text-guided image-to-image generation on WikiArt. Recall measures the diversity of the images and is only listed as a reference.}
   \label{table:generation_eval}
\end{table}

We also evaluate the quality of the adversarial example generated by different methods. As shown in Table \ref{table:image_quality}, the image quality of our adversarial examples is slightly higher than that of the existing methods. Moreover, the generation is significantly faster than the existing methods since we directly use the trained generator instead of iterative optimization against DMs. Our method has an advantage when generating a large number of examples.

\begin{table}[t]
   \begin{center}
   \begin{tabular}{ccccc}
   \hline
   Method & MSE $\downarrow$ & PSNR $\uparrow$ & SSIM $\uparrow$ & Run time \\ 
   \hline
   AdvDM  & 0.0038 & 29.1 & 0.80 & 32s \\
   Mist  & 0.0040 & 29.0 & \textbf{0.81} & 35s\\
   Ours  & \textbf{0.0037} & \textbf{30.1} & 0.80 & \textbf{0.2s}\\
   \hline
   \end{tabular}
   \end{center}
   \caption{The image quality and generation time (per image) of the adversarial examples on WikiArt.}
   \label{table:image_quality}
\end{table}

For qualitative evaluation, we visualize two examples generated by different methods in Figure \ref{fig:sample-eg}. More examples are shown in Appendix \ref{sec:img2img}. For the adversarial examples, all methods seem to introduce only subtle perturbations that are almost invisible to human eyes. For the generated images based on the adversarial examples, the existing methods succeed in adding chaotic textures to the generated image. However, such changes can be difficult to understand and evaluate. On the other hand, our adversarial examples succeed in generating images with visible watermarks as well as chaotic textures. Our method provides a more straightforward way to indicate copyright violations.

\begin{figure}[t]
   \begin{center}
      \includegraphics[width=0.99\linewidth]{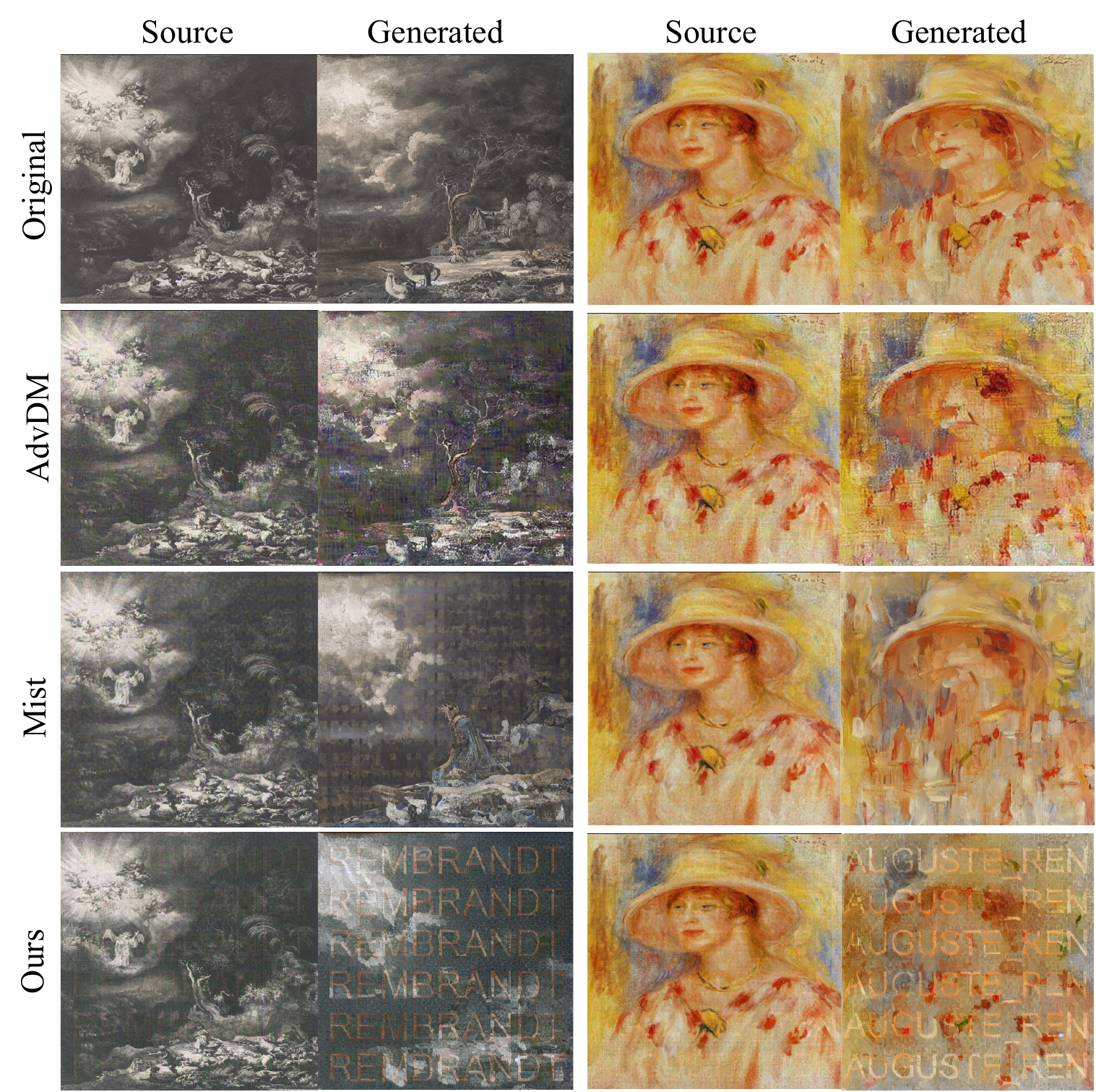}
   \end{center}
      \caption{Comparison of different methods under text-guided image-to-image generation. Source is the image input to the LDM, and generated image is the output of the LDM (strength 0.3). The watermarks of our method are designed according to the artist name (from left to right: REMBRANDT, AUGUSTE\_REN).}
   \label{fig:sample-eg}
\end{figure}

\subsection{Textual Inversion}
\label{sec:main_Textual Inversion}
Textual inversion is another important scenario related to copyright violations. In this scenario, given a small number of images, we first learn to represent the concept of the images in a new word $S_*$, then use $S_*$ to guide image generation. Following the method of textual inversion \cite{gal2022image}, we separate our evaluation images into 4-image groups for each artist or ImageNet category. For each 4-image group, we set the maximum steps to optimize $S_*$ to 5000. After the optimization, we generate 10 images for each $S_*$ under 10 prompt templates. This process is implemented for both original images and our adversarial examples.

For evaluation, we compare our method with existing methods, and the results are shown in Table \ref{table:textual_inversion}. For the generated image, our method has higher NCC and FID, which indicates our method successfully injects the watermark information in the generation of the adversarial example. In addition, our adversarial example also has higher PSNR and SSIM, which means the generated perturbation is subtle. An example of textual inversion is shown in Figure \ref{fig:textualinversion}, and examples comparing with the previous methods are shown in Appendix \ref{sec:textualinversion}. The features of the original images can be learned and used to guide image generation. On the other hand, when applying our method, images are generated with very clear watermarks, which can be useful for protecting copyrighted images.

\begin{table}[t]
   \begin{center}
   \begin{tabular}{c|cc|cc}
   \hline
   & \multicolumn{2}{c|}{Generated image} & \multicolumn{2}{c}{Adversarial example} \\
   Method & NCC $\uparrow$ & FID $\uparrow$ & PSNR $\uparrow$ & SSIM $\uparrow$ \\ 
   \hline
   Original & 0 & 120.7 & - & - \\
   AdvDM  & 0 & 327.3 & 26.3 & 0.73 \\
   Mist  & 0.15 & 318.4 & 26.7 & 0.74\\
   Ours  & \textbf{0.40} & \textbf{412.9} & \textbf{27.1} & \textbf{0.75} \\
   \hline
   \end{tabular}
   \end{center}
   \caption{The performance of the generated image and adversarial examples on ImageNet under textual inversion.}
   \label{table:textual_inversion}
\end{table}

\begin{figure*}[t]
   \begin{center}
      \includegraphics[width=0.935\linewidth]{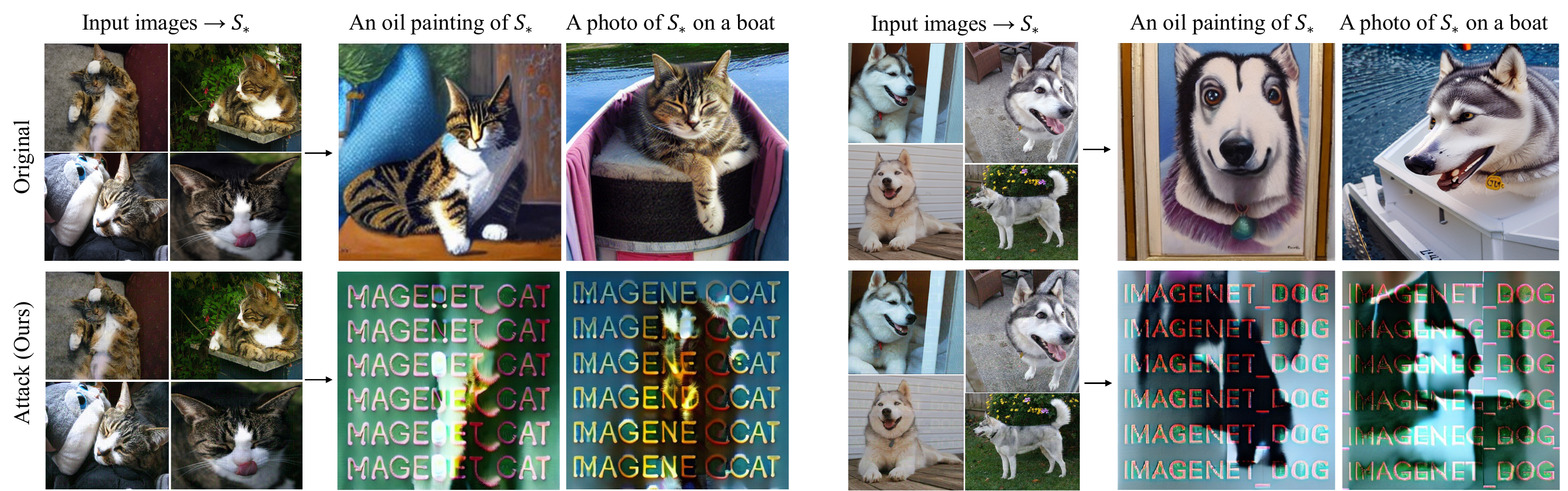}
   \end{center}
      \caption{Examples of textual inversion on ImageNet. The watermarks used in the examples are IMAGENET\_CAT and IMAGENET\_DOG.}
   \label{fig:textualinversion}
\end{figure*}

\subsection{Ablation Study}
\noindent
\textbf{Weights of the losses.} This experiment aims to study the mechanism of our method. In equation \ref {equation:total}, we design three losses to train the generator. As our target is to attack DMs, the adversarial loss $\mathcal{L}_{adv}$ is necessary, and we set its weight to 1. We then evaluate the effect of the GAN loss $\mathcal{L}_{GAN}$, the perturbation loss $\mathcal{L}_{pert}$, and the weight used to control the perturbation of the watermark region $w$ by setting their weights $\alpha$, $\beta$, and $w$ to 0 or their default value. 

The results are shown in Table \ref{table:weight_balance}. The first row is the result of only using $\mathcal{L}_{adv}$ and setting other losses to 0. In this case, although the attack ability for DMs is high, the generated adversarial examples are full of noise. By comparing the first and second rows, we observe that $\mathcal{L}_{GAN}$ significantly improves the image quality of the adversarial example, even though the attack ability decreases. On the other hand, $\mathcal{L}_{pert}$ which aims to bound the magnitude of the perturbation, also has an effect similar to that of $\mathcal{L}_{GAN}$ (as seen when comparing first and third rows). In addition, adding weight term to $\mathcal{L}_{pert}$ further improves the quality of the adversarial examples without sacrificing too much attack ability (as seen when comparing third and fourth rows). Finally, we combine all these terms to achieve a balanced performance between image quality and attack ability.

\begin{table}[t]
   \begin{center}
   \begin{tabular}{ccc|cc|cc}
   \hline
   \multicolumn{3}{c|}{Weights} & \multicolumn{2}{c|}{Generated image} & \multicolumn{2}{c}{Adversarial example} \\
   $\alpha$ & $\beta$ & $w$ & NCC $\uparrow$ & FID $\uparrow$ & PSNR $\uparrow$ & SSIM $\uparrow$ \\ 
   \hline
   $\times$ & $\times$  & $\times$ & 0.60 & 565.3 & 13.1 & 0.12 \\
   \checkmark & $\times$  & $\times$ & 0.41 & 350.6 & 22.0 &  0.65\\
   $\times$ & \checkmark  & $\times$  & 0.35 & 318.2 & 24.2 & 0.70 \\
   $\times$  & \checkmark & \checkmark & 0.34 & 292.2 & 27.8 & 0.76 \\
   \checkmark & \checkmark & \checkmark & 0.31 & 245.5 & 30.1 & 0.80 \\
   \hline
   \end{tabular}
   \end{center}
   \caption{The effect of the GAN loss ($\alpha$), the perturbation loss ($\beta$), and the weight of the watermark region ($w$) on generating adversarial examples (WikiArt) under image-to-image generation.}
   \label{table:weight_balance}
\end{table}

\noindent
\textbf{Maximum perturbation bound.}
In the case of adversarial examples, there is always a trade-off between image quality and the success rate of the attack. To explore this trade-off, we conducted experiments with different bounds. We used PSNR to measure the image quality of the adversarial examples, and NCC to assess the visibility of the watermark in the generated image which indicates the attack success rate. Table \ref{table:maxbound} shows that as the bound increases, the watermark on the generated image becomes more distinct. However, this may also cause the watermark on the adversarial image to become more visible, which can reduce the image quality. In real-world applications, users can adjust the bounds to balance between achieving high-quality adversarial images and ensuring watermark visibility to protect their work.

\begin{table}[t]
    \begin{center}
    \begin{tabular}{ccccccc}
    \hline
    Bound/255 & 2 & 6 & 10 & 15 & 20 \\ 
    \hline
    PSNR $\uparrow$ & 40.1 & 34.2 & 30.1 & 24.2 & 17.5\\
    NCC $\uparrow$ & 0.13 & 0.22 & 0.31 & 0.37 & 0.41 \\
    \hline
    \end{tabular}
    \end{center}
    \caption{The effect of using different perturbation bound. The experiment uses WikiArt under image-to-image generation.}
    \label{table:maxbound}
\end{table}

\noindent
\textbf{The number of samples for training.} To generate adversarial examples that contain a specific watermark, we need to include samples with that watermark in the training data. In this experiment, we analyze how many samples we need for a watermark to be learned by the generator. We choose a different number of samples to train a generator for a specific watermark. Please note that different watermarks can be trained together within the same generator, and this is the default setting for our other experiments. The results of training one personal watermark are in Table \ref{table:sample_num}. We observe there is no large performance difference between using 10 samples and 100 samples for training, which shows that our method only needs a small number of samples to train to embed a specific watermark. Therefore, one use case of our method is to create a personal watermark generator using a few samples within 2-3 min, and then this generator can be used to embed the personal watermark for any of that user's creations to protect copyright.

\begin{table}[t]
   \begin{center}
   \begin{tabular}{ccccc}
   \hline
   Samples & NCC $\uparrow$ & FID $\uparrow$ & prec. $\downarrow$ & train time \\  
   \hline
   1 & 0.12 & 177.8 & 0.32 & 20 sec\\
   5 & 0.28 & 235.3 & 0.04 & 1 min  \\
   10 & 0.30 & 240.6 & 0.03 & 2 min \\
   100 & 0.31 & 243.5 & 0.02 & 20 min \\
   \hline
   \end{tabular}
   \end{center}
   \caption{The effect and training time of using different number of samples to train a generator for a specific watermark. The experiment is conducted under image-to-image generation on WikiArt.}
   \label{table:sample_num}
\end{table}

\subsection{Robustness of adversarial examples}
To evaluate the robustness of the adversarial examples generated by our method, we apply several widely used adversarial defenses including JPEG compression \cite{das2018shield}, Randomized Smoothing (RS) \cite{cohen2019certified}, and Total Variance Minimization (TVM) \cite{guo2017countering}. Such defenses are directly processed on the adversarial examples to eliminate the perturbations. 

\begin{figure}[t]
   \begin{center}
      \includegraphics[width=1.0\linewidth]{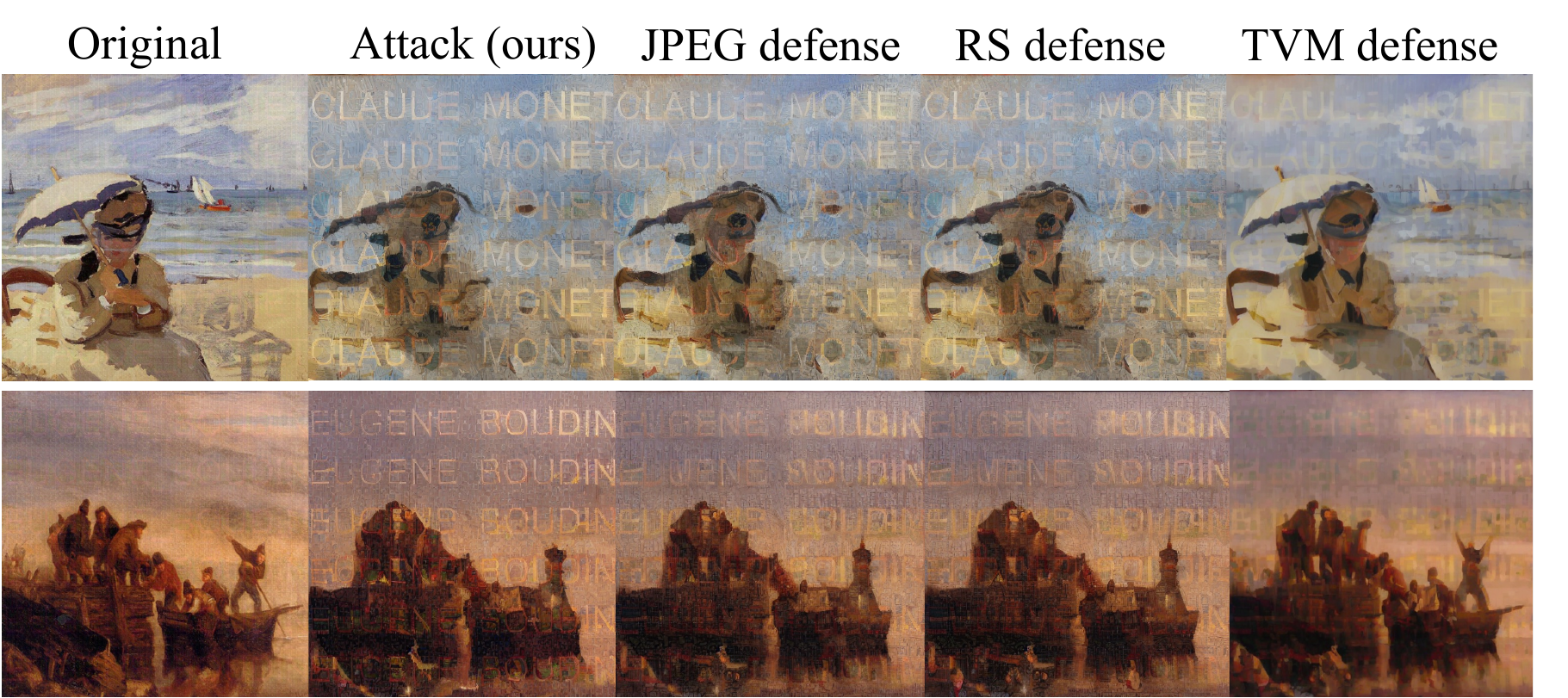}
   \end{center}
      \caption{Examples of the generated image from original image, attack image, and attack image with defense.}
   \label{fig:defense}
\end{figure}

The results of applying attacks and defenses are shown in Table \ref{table:defense}. The ``No attack” column means applying defense on the original image to generate images. There is no watermark being embedded so NCC almost equals 0, and applying defenses slightly increases FID since the quality of the original image decreases. On the other hand, the ``Attack” column means we first generate the adversarial examples, apply defenses to them (except the first row), and then use them to generate new images. We observe that NCC and FID both decrease when adding defenses to our adversarial examples. However, compared with ``No attack”, the ``Attack” column still has higher NCC and FID which means the defenses cannot completely eliminate the effect of our attack. In Figure \ref{fig:defense}, we show some examples under defenses, and the watermark is still visible for all cases.

\begin{table}[t]
   \begin{center}
   \begin{tabular}{c|cc|cc}
   \hline
   & \multicolumn{2}{c|}{No attack (Origin)} & \multicolumn{2}{c}{Attack (Ours)} \\
   Defense & NCC $\uparrow$ & FID $\uparrow$ & NCC $\uparrow$ & FID $\uparrow$ \\  
   \hline
   No defense & 0 & 120.7 & 0.31 & 245.5 \\
   JPEG & 0 & 124.1 & 0.21 & 202.2  \\
   RS & 0 & 123.2 & 0.20 & 195.6 \\
   TVM & 0 & 132.5 & 0.16 & 166.8  \\
   \hline
   \end{tabular}
   \end{center}
   \caption{The effects of applying defenses on adversarial examples under image-to-image generation.}
   \label{table:defense}
\end{table}

\subsection{Transferability on Other Generative Models}
In this section, we explore the transferability of our method on other generative models. This is also known as the black-box attack, where one target DM is used to train the generator and obtain the adversarial examples, and then these adversarial examples are directly used to attack other models. We generate adversarial examples on Stable Diffusion 1.5 (SD 1.5) \footnote{https://huggingface.co/runwayml/stable-diffusion-v1-5} and test on other models or tools including Dreamshaper 8 \footnote{https://civitai.com/models/4384/dreamshaper} , NovelAI \footnote{https://novelai.net/}, and Runway AI magic tools (image variation) \footnote{https://runwayml.com/ai-magic-tools/image-to-image/} under image-to-image generation. The details of the settings are described in the Appendix \ref{sec:settings_of_other_models}. 

The results are shown in Figure \ref{fig:black_box}, and more examples can be found in Appendix \ref{sec:transferability}. Compared to images generated from the original image (first row), the images generated from our adversarial examples (second row) contain chaotic textures and a visible watermark for most cases. For the result of the Runway tool, although there is no obvious watermark on the generated image (as no parameter can be adjusted), our method still generates an image that differs significantly from the original one. Even though the watermarks are not as strong as they are in the white-box setting, our method still demonstrates good transferability. These results show that our method has the potential to prevent copyright violations from various image generative models.

\begin{figure}[t]
   \begin{center}
      \includegraphics[width=0.999\linewidth]{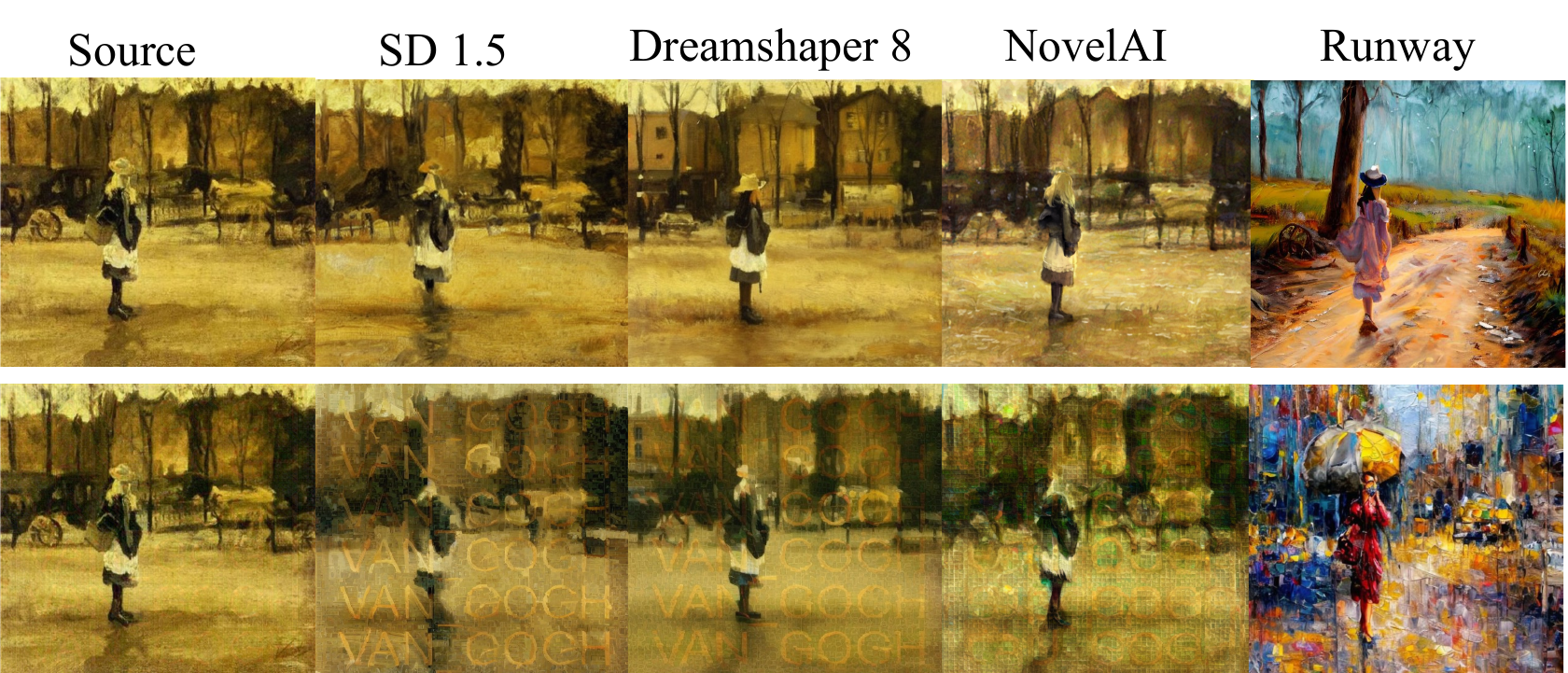}
   \end{center}
      \caption{Black-box attack to other models under image-to-image generation. The first row: generated images from the original image. Second row: generated images from our adversarial example.}
   \label{fig:black_box}
\end{figure}

%% file: sec/5_conclusions.tex
\section{Conclusion}
In this work, we propose a novel method for copyright prevention against DMs. We build a generator to embed personal watermarks into the adversarial examples, and such examples can force DMs to generate images with visible watermarks and chaotic textures. We use a conditional GAN architecture with three carefully designed losses to optimize the generator. Experiments show our method performs well in various image generation scenarios and exhibits good transferability to attack other models. Additionally, our generation is significantly faster than those of previous methods. Therefore, this work provides a simple yet powerful way to protect image copyright against DMs.

%% file: sec/X_suppl.tex
\clearpage
\setcounter{page}{1}
\maketitlesupplementary

\renewcommand{\thesection}{A}
\section{Image Generation under Other Scenarios}
\label{sec:OtherScenarios}
\subsection{DreamBooth}
In this experiment, we evaluate the performance of our method on DreamBooth \cite{ruiz2023dreambooth} which is another method to personalize text-to-image models.  Given a few (3-5) images of a subject, DreamBooth fine-tunes the pre-trained text-to-image model to learn a unique identifier for that subject. After the fine-tuning, we can use the unique identifier to generate contextualized images of the subject in different scenes, poses, and views. We perform the DreamBooth fine-tuning by using the Python library diffusers \footnote{https://github.com/huggingface/diffusers} on both the original images and adversarial examples. We set the resolution to 512, the learning rate to $2\times10^{-6}$ and the maximum number of train steps to 2000. The steps and parameters used to generate our adversarial example are the same as those used in other image generations in Section \ref{sec:imagegen_settings}.

\begin{figure*}[t]
   \begin{center}
      \includegraphics[width=0.99\linewidth]{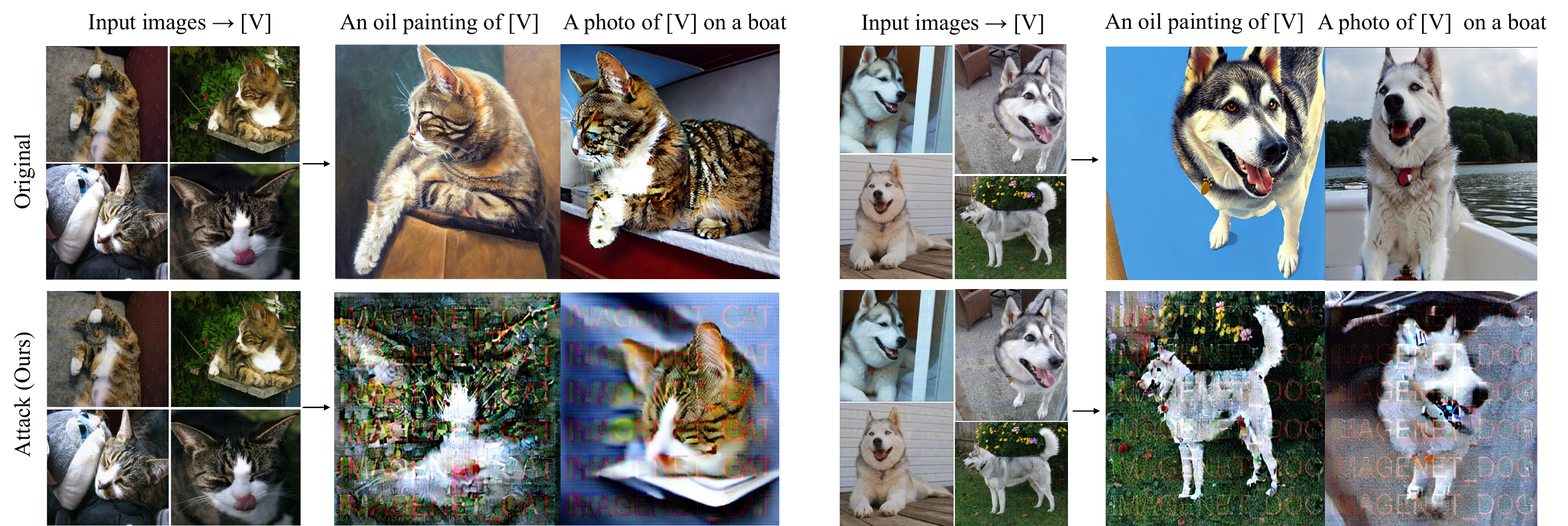}
   \end{center}
      \caption{Comparison between the original result and our attack result under DreamBooth on ImageNet. The watermarks used in the examples are IMAGENET\_CAT and IMAGENET\_DOG.}
   \label{fig:dreambooth}
\end{figure*}

Visualization examples are shown in Figure \ref{fig:dreambooth}. We compare the results generated from the original images with those from our adversarial examples. For the original images, the features of the input images can be learned and used to generate new images. On the other hand, after applying our attack, images are generated with visible watermarks and chaotic textures. Therefore, our method is also applicable to DreamBooth settings to protect copyrights.

\subsection{LoRA Fine-tuning}
Low-Rank Adaptation (LoRA) \cite{hu2021lora} is a method initially proposed for fine-tuning large-language models. It freezes pre-trained model weights and injects trainable layers in each transformer block. In the case of fine-tuning diffusion models, LoRA can be applied to the cross-attention layers that relate the image representations with their corresponding prompts. We perform the LoRA fine-tuning by using the Python library diffusers on both the original images and our adversarial examples. We set the resolution to 512, the learning rate to $10^{-4}$, the maximum gradient norm to 1.0 and the maximum number of train steps to 2000.

\begin{figure*}[t]
   \begin{center}
      \includegraphics[width=0.99\linewidth]{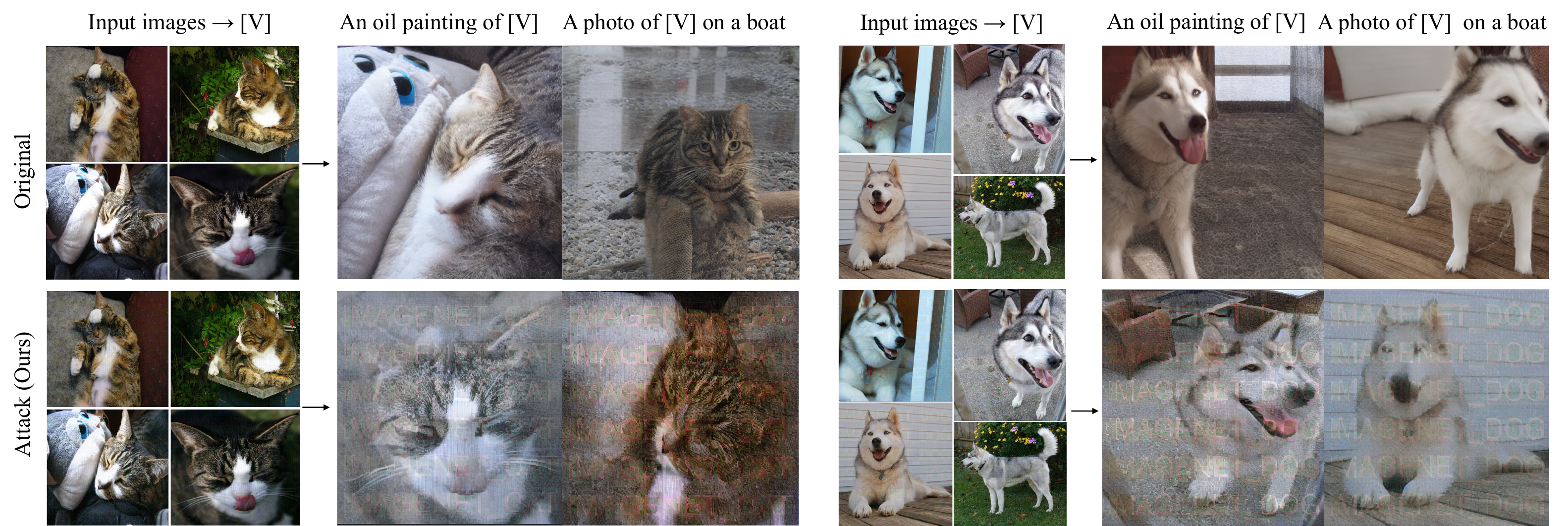}
   \end{center}
      \caption{Comparison between the original result and our attack result under LoRA fine-tuning on ImageNet. The watermarks used in the examples are IMAGENET\_CAT and IMAGENET\_DOG.}
   \label{fig:lora}
\end{figure*}

Visualization examples are presented in Figure \ref{fig:lora}. We compare the results generated from the original images with those generated using our adversarial examples. For the original images, although the effectiveness of LoRA fine-tuning is not as pronounced as that of textual inversion and Dreambooth, the features of the input images can still be extracted to generate new images. On the other hand, after applying our attack, the resulting images exhibit visible watermarks and chaotic textures. Consequently, our method could be effective in LoRA fine-tuning settings for copyright protection purposes.

\subsection{Custom Diffusion}
Custom Diffusion \cite{choi2023custom} is another fine-tuning technique for personalizing image generation models. Given a few (~4-5) example images, Custom Diffusion works by only training weights in the cross-attention layers, and it uses a special word to represent the newly learned concept. We perform the Custom Diffusion by using the Python library diffusers on both the original images and our adversarial examples. We use single-concept fine-tuning and set the resolution to 512, the learning rate to $10^{-5}$ and the maximum number of train steps to 250.

\begin{figure*}[t]
   \begin{center}
      \includegraphics[width=0.99\linewidth]{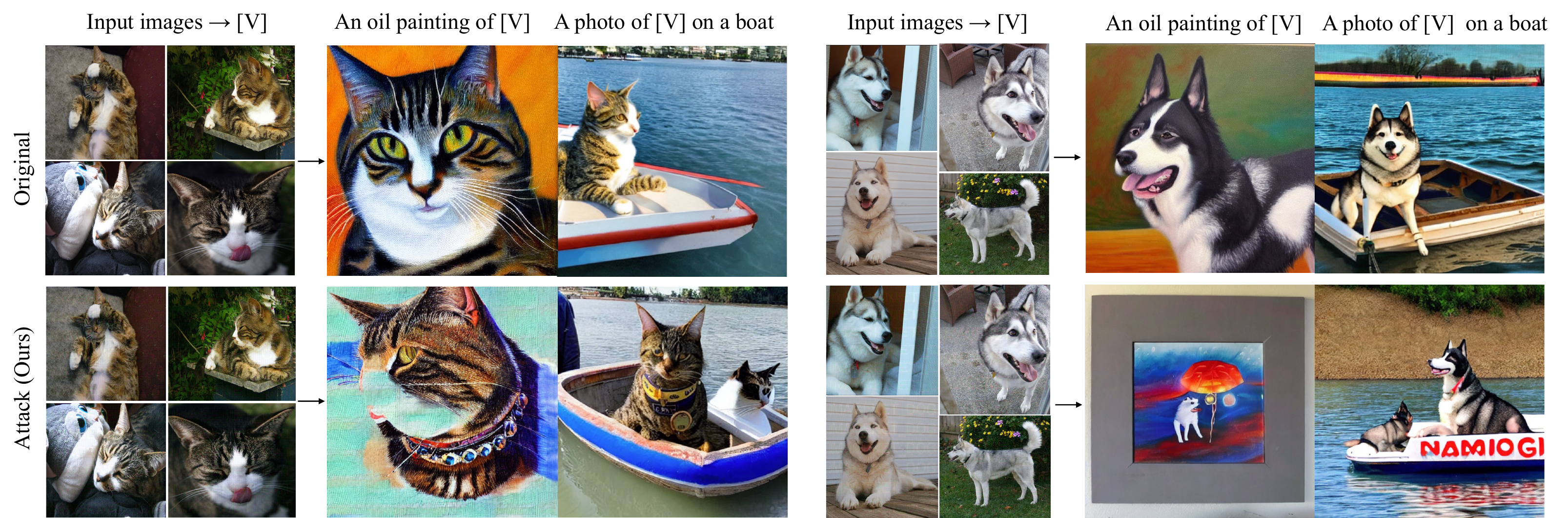}
   \end{center}
      \caption{Comparison between the original result and our attack result under Custom Diffusion on ImageNet. The watermarks used in the examples are IMAGENET\_CAT and IMAGENET\_DOG.}
   \label{fig:customdiffusion}
\end{figure*}

Visualization examples are presented in Figure \ref{fig:customdiffusion}. We compare the results generated from the original images with those generated using our adversarial examples. For the original images, the features can be learned and utilized to generate new images. In the case of the adversarial examples, although no explicit watermark is visible on the generated images, many generated images contain irrelevant text, and many generated content appears distorted and unnatural. Despite the outcome being different from our intended target, our method still demonstrates potential for preventing image imitation under the Custom diffusion.

\renewcommand{\thesection}{B}
\section{Implementation Details of the Model}
\label{sec:modeldetails}
\textbf{Generator architecture} We adopt the naming conventions established by \cite{johnson2016perceptual, zhu2017unpaired}. Here, $dk$ refers to a $3\times3$ Convolution-InstanceNorm-ReLU layer with $k$ filters and stride 2. The notation $Rk$ represents a residual block that contains two $3\times3$ convolution layers with the same number of $k$ filters. Lastly, $uk$ signifies a $3\times3$ fractional-strided-Convolution-InstanceNorm-ReLU layer with $k$ filters and a stride of 1/2.

The generator is composed of the following layers:
\setlist{nolistsep} 
\begin{itemize}
\item The encoder consists of a series of downsampling layers: $d64, d128, d256$.
\item This is followed by a series of residual blocks: $R256, R256, R256, R256$.
\item The decoder then upsamples the feature maps: $u128, u64$.
\item The final layer of the generator is a $3\times3$ Convolution-Tanh layer with the number of output channels corresponding to the image's number of channels.
\end{itemize}

\noindent
\textbf{Discriminator architecture} We utilize a convolutional neural network (CNN) structure inspired by \cite{radford2015unsupervised}. The notation $Ck$ indicates a $4\times4$ Convolution-LeakyReLU layer with $k$ filters and stride 2. We do not apply Instance Normalization to the first layer of the discriminator. Leaky ReLUs are used with a negative slope of 0.2.

The discriminator is composed of the following layers:
\setlist{nolistsep} 
\begin{itemize}
\item A series of convolutional layers with increasing filter sizes: $C64, C128, C256, C512, C1024$.
\item The final convolutional layer is a $16\times16$ Convolution layer that outputs a single feature map.
\item A Sigmoid activation function is applied to the output of the last layer to obtain a probability value indicating whether the input image is original or adversarial.
\end{itemize}

\renewcommand{\thesection}{C}
\section{More Results of Text-guided Image-to-image Generation}

\subsection{Results under Different Strength Value}
\label{sec:strengthresult}
\begin{figure}[t]
   \begin{center}
      \includegraphics[width=0.98\linewidth]{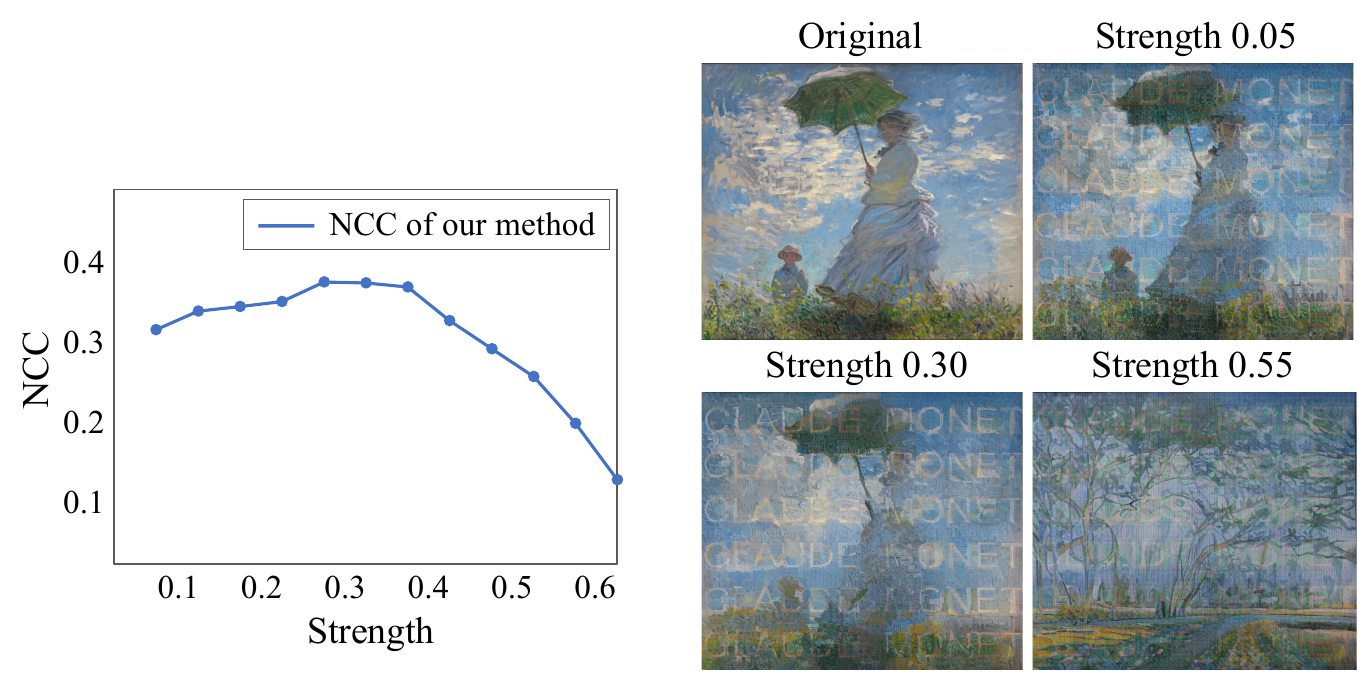}
   \end{center}
      \caption{The influence of strength parameter on the visibility of watermark under image-to-image generation.}
   \label{fig:strength}
\end{figure}

The strength parameter plays a crucial role in image-to-image generation. It determines the level of noise that is added to the original image while generating new images. A small strength value will produce an image nearly identical to the original, while a large strength value will produce an image that largely differs from the original. We explore the change in the watermark on the generated images with different strength values. As shown in Figure \ref{fig:strength}, NCC remains at a high value when the strength is smaller than 0.4 and drops after that. We also observe that when the strength is increased to 0.55, the content of the generated image is already far from the original one, which makes it difficult to identify as a copyright violation. Therefore, we focus on the setting with a strength smaller than 0.5, and our method can successfully attack DMs under this setting.

\subsection{Results under Varied Prompts}
\label{sec:variedprompts}
For text-guided image-to-image generation, the ideal approach is to use varied prompts to guide the generation of new images. However, with thousands of diverse test images, preparing individual prompts for each image is time-consuming. Consequently, we used a uniform prompt for both training and inference. 

On the other hand, to explore the influence of different prompts, we evaluated 40 images with 5 varied prompts per image. The Normalized Cross-Correlation (NCC) results between the generated images and the watermark were similar for both fixed prompts (0.31) and varied prompts (0.30). This similarity may be due to the strength parameter being the primary determinant of the level of noise added, while the prompts merely guide the noise during generation. Therefore, even with varied prompts, the watermarks on the generated images remain almost unchanged when the strength parameter is constant.

\renewcommand{\thesection}{D}
\section{More Visualization Examples}

\subsection{Text-guided Image-to-Image Generation}
\label{sec:img2img}
\begin{figure*}[t]
   \begin{center}
      \includegraphics[width=0.95\linewidth]{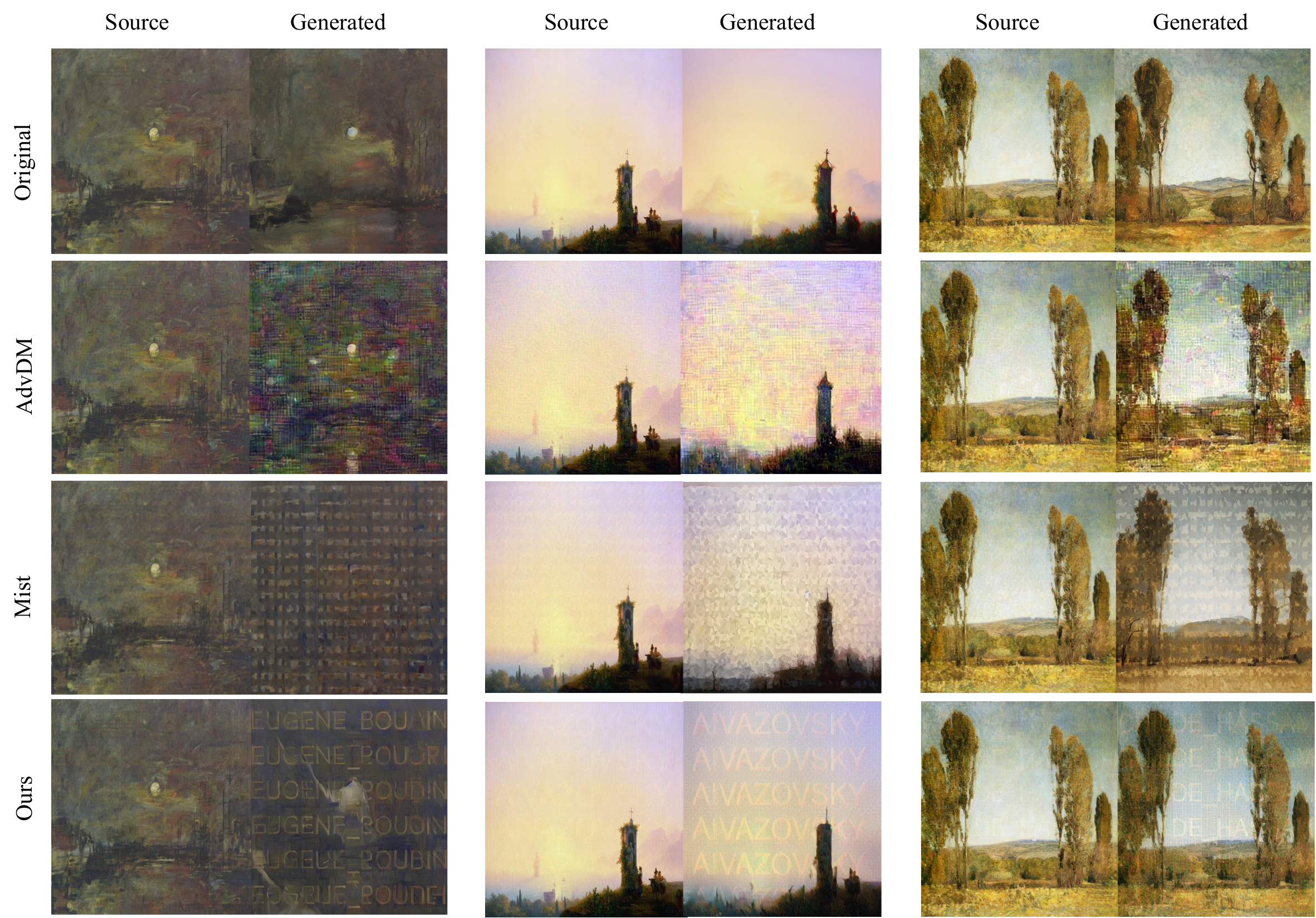}
   \end{center}
      \caption{More examples of text-guided image-to-image generation on WikiArt.}
   \label{fig:appen_img2img}
\end{figure*}

We show more examples under image-to-image generation in Figure \ref{fig:appen_img2img}. The experimental settings remain consistent with those in Section \ref{sec:Text-guided Image-to-Image Generation}. Compared to the existing methods that only add chaotic content, our method adds visible watermarks, providing a more straightforward way to show copyright violations.

\subsection{Textual Inversion}
\label{sec:textualinversion}

\begin{figure*}[t]
   \begin{center}
      \includegraphics[width=0.92\linewidth]{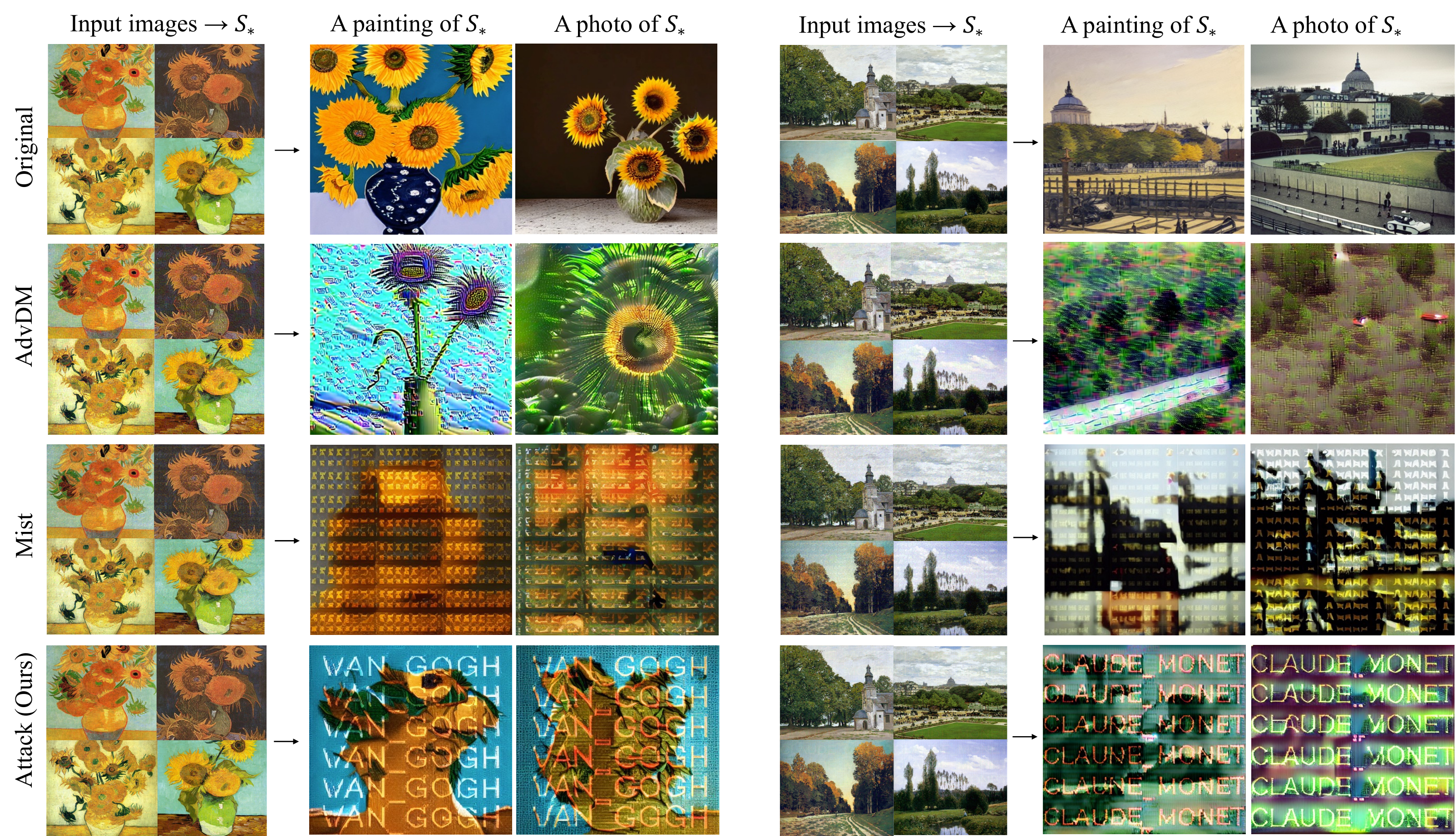}
   \end{center}
      \caption{Comparison between our method and the previous methods under textual inversion on WikiArt.}
   \label{fig:appen_textualinv}
\end{figure*}

We show the comparison result between our method and the previous methods under textual inversion in Figure \ref{fig:appen_textualinv}. The experimental settings remain consistent with those in Section \ref{sec:main_Textual Inversion}. The previous methods could encourage DMs to generate images that are far from the original ones (AdvDM) or with large artifacts (Mist). However, such changes are not straightforward enough to indicate copyright violations. Our method succeeds in compelling DMs to generate images with obvious watermarks, demonstrating a simple yet powerful way to prevent copyright violations.

\subsection{Transferability on Other Generative Models}
\label{sec:transferability}
\begin{figure*}[t]
   \begin{center}
      \includegraphics[width=0.85\linewidth]{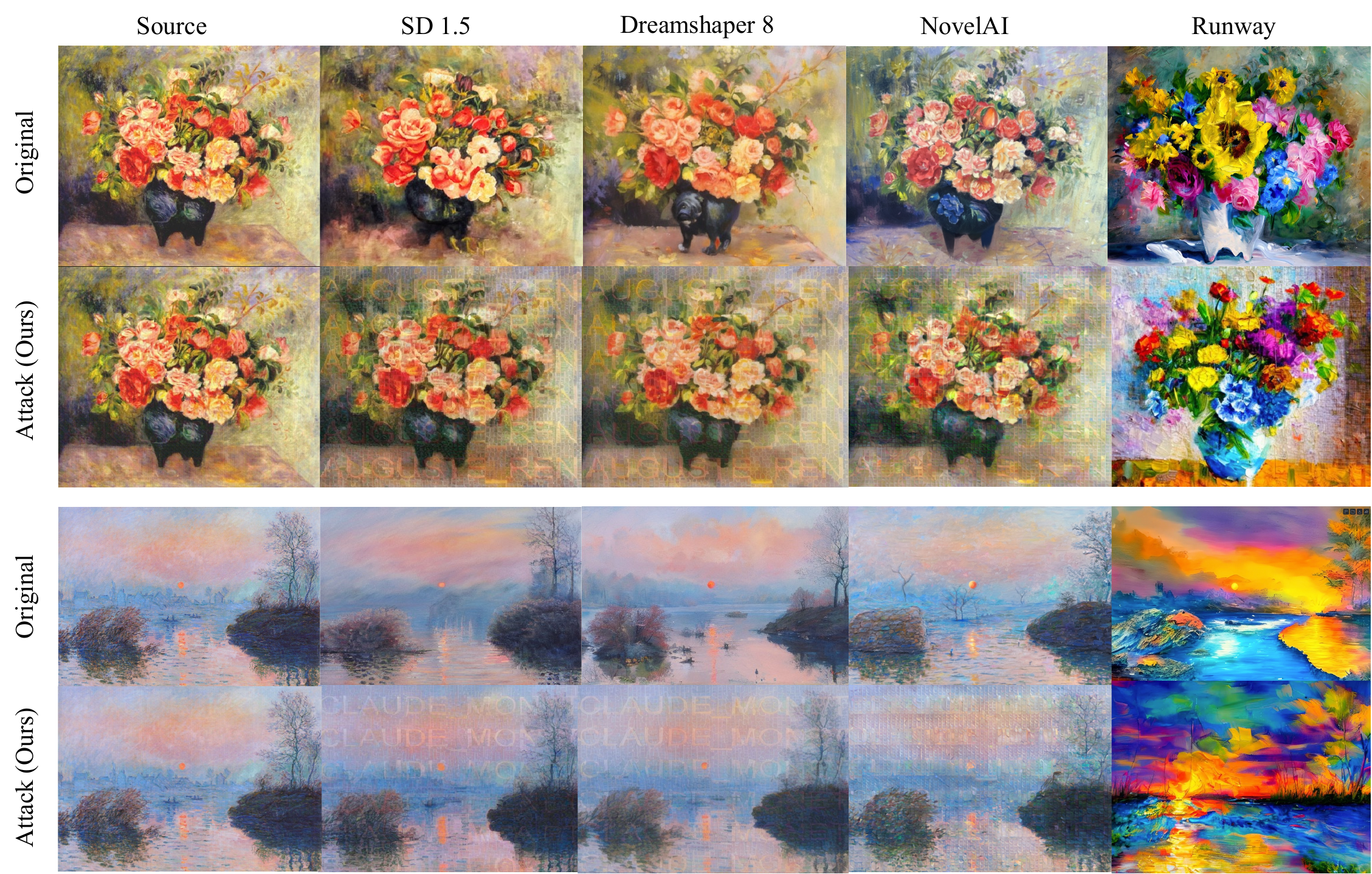}
   \end{center}
      \caption{More examples of black-box attack under image-to-image generation using various models or tools.}
   \label{fig:appen_black-box}
\end{figure*}

 We show more visualization examples of this setting in Figure \ref{fig:appen_black-box}. All examples follow a similar trend in that our method can add visible watermarks to the generated images for most models. For Runway, although there is no obvious watermark, our method still forces the models to generate images that differ from the original ones. Our method exhibits good transferability and can attack various image-generation models.

\renewcommand{\thesection}{E}
\section{Settings of Other Generative Models}
\label{sec:settings_of_other_models}
\noindent
\textbf{Stable Diffusion 1.5 (SD 1.5) \footnote{https://huggingface.co/runwayml/stable-diffusion-v1-5}} is a latent text-to-image diffusion model capable of generating realistic images. It can also be applied to image-to-image generation by passing a text prompt and an initial image to condition the generation of new images. We use the text prompt ``A painting" for the WikiArt dataset, and ``A photo" for the ImageNet dataset. We set the sampling methods to DPM++ 2M Karras, the sampling steps to 50, and the strength to 0.30 as default. Please note that this is the model we used to generate our adversarial examples.

\noindent
\textbf{Dreamshaper 8 \footnote{https://civitai.com/models/4384/dreamshaper}} is a fine-tuned version of Stable Diffusion that addresses some of its limitations. It improves the convergence speed, handles high-dimensional data, and is more robust to noise. We conduct the experiments under image-to-image generation and set the sampling method to DPM++ 2M Karras, the steps to 40, the guidance to 10, and the strength to 0.30 as default.

\noindent
\textbf{NovelAI \footnote{https://novelai.net/}} is a service that creates unique stories with virtual companionship. It also has the potential to raise concerns about copyright violations. We conduct experiments under the image-to-image generation scenario in NovelAI. We use the text prompt ``A painting" for the WikiArt dataset, and ``A photo" for the ImageNet dataset. We set the resolution to 512, the sampling method to DPM++ 2M, the steps to 40, the guidance to 10, and the strength to 0.30 as default.

\noindent
\textbf{Runway AI magic tools \footnote{https://runwayml.com/ai-magic-tools/image-to-image/}} provides various creative tools to ideate, generate and edit images and videos. We conduct experiments under image variation in this tool. Since there is no parameter that can be adjusted, we directly input our images into this tool and obtain the generated image.